\pdfoutput=1
\documentclass[11pt,a4paper]{article}
\usepackage[hyperref]{emnlp2018}
\usepackage{times}
\usepackage{latexsym}

\usepackage{url}

\usepackage{amsmath}
\usepackage{float}
\usepackage{multirow}
\usepackage{tikz}

\aclfinalcopy % Uncomment this line for the final submission

\title{Incremental processing of noisy user utterances in the spoken language understanding task}

\author{Stefan Constantin$^*$ ~~~~~ Jan Niehues$^+$ ~~~~~ Alex Waibel$^*$ \\
{$^*$ Karlsruhe Institute of Technology} \\
{Institute for Anthropomatics and Robotic} \\
{\tt $\{$stefan.constantin|waibel$\}$@kit.edu} \\
\\
{$^+$ Maastricht University} \\
{Department of Data Science and Knowledge Engineering} \\
{\tt jan.niehues@maastrichtuniversity.nl}
}

\date{}

\begin{document}
\maketitle

\begin{abstract}
The state-of-the-art neural network architectures make it possible to create spoken language understanding systems with high quality and fast processing time.
One major challenge for real-world applications is the high latency of these systems caused by triggered actions with high executions times.
If an action can be separated into subactions, the reaction time of the systems can be improved through incremental processing of the user utterance and starting subactions while the utterance is still being uttered.
In this work, we present a model-agnostic method to achieve high quality in processing incrementally produced partial utterances.
Based on clean and noisy versions of the ATIS dataset, we show how to create datasets with our method to create low-latency natural language understanding components.
We get improvements of up to 47.91 absolute percentage points in the metric F\textsubscript{1}-score.
\end{abstract}

\section{Introduction}
\label{sec:Introduction}
Dialog Systems are ubiquitous - they are used in customer hotlines, at home (Amazon Alexa, Apple Siri, Google Home, etc.), in cars, in robots \cite{Asfour2018}, and in smartphones (Apple Siri, Google Assistant, etc.).
From a user experience point of view, one of the main challenges of state-of-the-art dialog systems is the slow reaction of the assistants.
Usually, these dialog systems wait for the completion of a user utterance and afterwards process the utterance.
The processed utterance can trigger a suitable action, e.\,g. ask for clarification, book a certain flight, or bring an object.
Actions can have a high execution time, due to which the dialog systems react slowly.
If an action can be separated into subactions, the reaction time of the dialog system can be improved through incremental processing of the user utterance and starting subactions while the utterance is still being uttered.
The action still has the same execution time but the action is completed earlier because it was started earlier and therefore the dialog system can react faster.
In the domain of airplane travel information, database queries can be finished earlier if the system can execute subqueries before the completion of the user utterance, e.\,g. the utterance \textit{On next Wednesday flight from Kansas City to Chicago should arrive in Chicago around 7 pm} can be separated in the databases queries \textit{flight from Kansas City to Chicago on next Wednesday} and \textit{use result of the first query to find flights that arrive in Chicago around 7 pm}.
In the domain of household robots, e.\,g. the user goal of the user utterance \textit{Bring me from the kitchen the cup that I like because it reminds me of my unforgettable vacation in the United States} can be fulfilled faster if the robot goes to the kitchen before the user utters what object the robot should bring.

Motivated by this approach to improve the reaction of dialog systems, our main contribution is a low-latency natural language understanding (NLU) component.
We use the Transformer architecture \cite{VaswaniSPUJGKP2017} to build this low-latency NLU component, but the ingredient to understand partial utterances and incrementally process user utterances is the model-agnostic training process presented in this work.
Secondly, partial utterances are particularly affected by noise.
This is due to the short context available in partial utterances and because automatic speech recognition (ASR) systems cannot utilize their complete language model and therefore potentially make more errors when transcribing short utterances.
We address the potential noisier inputs by including noisy inputs in the training process.
Finally, we present two evaluation schemes for low-latency NLU components.

\section{Related Work}
\citet{GambinoZKS2018} described time buying strategies to avoid long pauses, e.\,g. by uttering an acknowledgement or echoing the user input.
However, the triggered actions are not finished earlier with this approach, but in cases where long pauses cannot be avoided, even with incremental processing, such time buying strategies can be applied.

The automatically generated backchannel described by \citet{RuedeMSW2017} gives feedback during the uttering of an utterance.
However, only acoustic features are used and it does not reduce the latency of actions that can be triggered by the utterances.

Studies have been conducted on incremental NLU.
\citet{DevaultST2009} used a maximum entropy classificator \cite{BergerPP1996} to classify the partial utterances.
They optimized the maximum entropy classificator for partial utterances by using an individual classificator for every utterance length.
The problem of this classification approach is that it is not suitable for tasks with a lot of different parameter combinations; for such tasks, a slot filling (sequence labeling task) or word by word approach (sequence to sequence task) is more suitable.
Such a more suitable approach is described by \citet{NiehuesPHSW2018} for incrementally updating machine translations.
The authors used an attention-based encoder decoder \cite{BahdanauCB2015}, which outputs a sequence.
We described and evaluated in this work such a more suitable approach for incremental NLU.

Different approaches are available to handle noisy input, such as general-purpose regularization techniques like dropout \cite{SrivastavaHKSS2014} and domain-specific regularization techniques e.\,g. data augmentation by inserting, deleting, and substituting words \cite{SperberNW2017}.
Our trained models in this work uses the general-purpose techniques and some of our trained models are trained with such augmented data to have a better performance on noisy data.

\section{Low-latency NLU component}
In this work, we present a model-agnostic method to build an incremental processing low-latency NLU component.
The advantages of this model-agnostic method are that we can use state-of-the-art neural network architectures and reuse the method for future state-of-the-art neural network architectures.
Our used architecture is described in Section \ref{sec:Architecture} and the used data is described in Section \ref{sec:Data}.
Our method to include the information necessary to incrementally process user utterances with high quality in the training dataset is described in Section \ref{sec:AdaptedDataPartial} and our methods to include noise to process noisy texts with high quality are described in Section \ref{sec:AdaptedDataNoise}.
In Sections \ref{sec:EvaluationMetrics} and \ref{sec:EvaluationData}, we present our evaluation metrics and evaluation schemes respectively.
The configuration of the used architecture is given in Section \ref{sec:Setup}.

\subsection{Architecture}
\label{sec:Architecture}
\begin{figure*}

\begin{tikzpicture}
\node[text width=1.4cm] (T1) at (0,2) {atis\_flight};
\node[text width=1.4cm] (T2) at (6,2) {atis\_flight};
\node[text width=4.1cm] (T3) at (12,2) {atis\_flight toloc pittsburgh};

\node[draw,rectangle] (M1) at (0,1){model};
\node[draw,rectangle] (M2) at (6,1){model};
\node[draw,rectangle] (M3) at (12,1){model};

\node[text width=1cm] (S1) at (0,0) {Flights};
\node[text width=1.5cm] (S2) at (6,0) {Flights to};
\node[text width=3.5cm] (S3) at (12,0) {Flights to Pittsburgh};

\draw [->] (S1) edge (M1);
\draw [->] (S2) edge (M2);
\draw [->] (S3) edge (M3);

\draw [->] (M1) edge (T1);
\draw [->] (M2) edge (T2);
\draw [->] (M3) edge (T3);
\end{tikzpicture}

\caption{model-agnostic approach}
\label{fig:ModelAgnosticApproach}
\end{figure*}
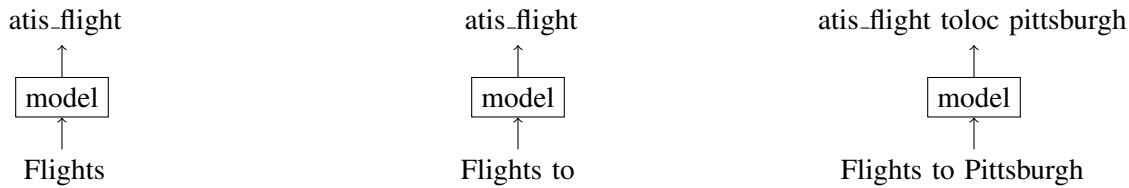

We used the Transformer architecture in our experiments to demonstrate the model-agnostic method.
The Transformer architecture, with its encoder and decoder, was used as sequence to sequence architecture.
The user utterances are the input sequences and their corresponding triggered actions are the output actions (this is described in more details in Section \ref{sec:Data}).
We used the Transformer implementation used by \citet{PhamNNMW2019} and added the functionality for online translation.
The original code\footnote{\url{https://github.com/quanpn90/NMTGMinor/tree/DbMajor}} and the added code are publicly available\footnote{\url{https://github.com/msc42/NMTGMinor/tree/DbMajor}}.
The partial utterances and, in the end, the full utterance were fed successively and completely into the Transformer architecture without using information of the computation of the previous partial utterances.
Our proposed method is model-agnostic because of this separate treatment and therefore an arbitrary model that can process sequences can be used to process the partial and full utterances.
The method is depicted in Figure \ref{fig:ModelAgnosticApproach} for the utterance \textit{Flights to Pittsburgh}.

\subsection{Data}
\label{sec:Data}
For our experiments, we used utterances from the Airline Travel Information System (ATIS) datasets.
We used the utterances that are used by \citet{HakkaniTCCGDW2016} and are publicly available\footnote{\url{https://github.com/yvchen/JointSLU}}.
These utterances were cleaned and every utterance is labeled with its intents and for every token, the corresponding slot is labeled with a tag (in the IOB2 format \cite{SangV1999} that is depicted in Figure \ref{fig:ATISS2S}).

\begin{figure*}
\centering

\begin{tabular}{ | p{2.7cm} | c | c | c | c | c | c | c | c | }
\hline
utterance \mbox{(source sequence)} & Which & flights & go & from & New & York & to & Pittsburgh \\
\hline
slots & O & O & O & O & B-fromloc & I-fromloc & O & B-toloc \\
\hline
intents & \multicolumn{8}{c |}{atis\_flight} \\
\hline
target~sequence & \multicolumn{8}{c |}{atis\_flight fromloc new york toloc pittsburgh} \\
\hline
\end{tabular}

\caption{joint intents classification and slot filling approach to end-to-end target sequence}
\label{fig:ATISS2S}
\end{figure*}

We converted the data from the IOB2 format to a sequence to sequence format \cite{ConstantinNW2019}.
The source sequence is a user utterance and the target sequence consists of the intents followed by the parameters.
In this work, the slot tag and the corresponding slot tokens compose an intents parameter.
An example of the conversion of the IOB2 format to the sequence to sequence format is depicted in Figure \ref{fig:ATISS2S}.
The sequence to sequence format has the advantages that no rules are needed for mapping the slot tokens to an API call or a database query and that this format is more robust against noisy text like \textit{What is restriction ap slash fifty seven}, where the noise word slash is introduced (in the classical IOB2 format, the tokens \textit{ap} and \textit{fifty seven} would not belong to the same chunk).

The publicly available utterances are partitioned in a training and test dataset.
The training dataset is partitioned in a training (train-2) and validation (dev-2) dataset.
Hereinafter, original training dataset refers to the utterances of the training dataset, training dataset refers to the utterances of the train-2 dataset, and validation dataset refers to the utterances of the dev-2 dataset.
We created a file that maps to every utterance in the training dataset the line number of the corresponding utterance in the original training dataset and a file that maps to every utterance in the validation dataset the line number of the corresponding utterance in the original training dataset.
We published these two files\footnote{\label{note:Msc42AtisData}\url{https://github.com/msc42/ATIS-data}}.
The training dataset has 4478 utterances, the validation dataset has 500 utterances, and the test dataset has 893 utterances.

The utterances were taken from the ATIS2 dataset (Linguistic Data Consortium (LDC) catalog number LDC93S5), the ATIS3 training dataset (LDC94S19) and the ATIS3 test dataset (LDC94S26).
The audio files of the spoken utterances and the uncleaned human transcribed transcripts are on the corresponding LDC CDs.
For the original training dataset and the test dataset, we published\footnote{see footnote \ref{note:Msc42AtisData}} in each case a file that maps to every utterance the path of the corresponding audio file and a file that maps to every utterance the path of the corresponding transcript of the corresponding LDC CD.
One audio file is missing on the corresponding LDC CD (LDC94S19): atis3/17\_2.1/atis3/sp\_trn/sri/tx0/2/tx0022ss.wav (corresponding to the training dataset).
We used the tool sph2pipe\footnote{\url{https://www.ldc.upenn.edu/sites/www.ldc.upenn.edu/files/ctools/sph2pipe_v2.5.tar.gz}} to convert the SPH files (with extension .wav) of the LDC CDs to WAVE files.

The utterances have an average token length of 11.21 - 11.36 in the training dataset, 11.48 in the validation dataset, and 10.30 in the test dataset.
We tokenized the utterances with the default English word tokenizer of the Natural Language Toolkit (NLTK)\footnote{https://www.nltk.org/} \cite{BirdKL2009}.

There are 19 unique intents in the ATIS data.
In the training dataset, 22 utterances are labeled with 2 intents and 1 utterance is labeled with 3 intents, in the validation dataset, there are 3 utterances with 2 intents and in the test dataset, there are 15 utterances with 2 intents, the rest of the utterances are labeled with 1 intent.
The intents are separated by the number sign in the target sequence.
The intents are unbalanced (more than 70\,\% of the utterances have the same intent, more than 90\,\% of the utterances belong to the 5 most used intents).
More information about the intents distribution is given Table \ref{fig:Intents_distribution}.
There are 83 different parameters that can parameterize the intents.
On average, a target has 3.35 (training dataset), 3.46 (validation dataset), and 3.19 (test dataset) parameters.

\subsection{Training process to improve incremental processing}
\label{sec:AdaptedDataPartial}
We call our dataset, which contains the dataset described in Section \ref{sec:Data}, \textit{cleaned full transcripts}.
Our model-agnostic method to achieve good quality for partial utterances works in this manner:
We use the dataset with the full utterances and create partial utterances from it.
An utterance of the length \(n\) is split into \(n\) utterances, where the \(i\)-th utterance of these utterances has the length \(i\).
The target contains all information that can be gotten from the source utterance of the length \(i\).
When only a part of a chunk is in the user utterance, only this part is integrated in the target utterances, e.\,g. \textit{I want a flight from New York to San} has the target \textit{atis\_flight fromloc.city\_name new york toloc.city\_name san}.
Such partial information contains information and can accelerate database queries, for example.
We created with this method the \textit{cleaned incremental transcripts} dataset.
An arbitrary model without modifications, in this work the Transformer architecture, can be trained with this dataset to have an improved incremental processing ability compared to a model trained only with full utterances.
Since every partial utterance is regarded as independent utterance, like the full utterances, our approach is model-agnostic.
The model-agnostic approach for the utterance \textit{Flights to Pittsburgh} is depicted in Figure \ref{fig:ModelAgnosticApproach}.

\subsection{Training process to improve robustness}
\label{sec:AdaptedDataNoise}
In Section \ref{sec:AdaptedDataPartial}, the training process for improving the incremental processing is described.
However, the described process does not consider the fact that the incremental data are noisier.
We induced noise in the training by training with artificial noise, human transcribed utterances that contain the noise of spoken utterances, and utterances transcribed by an ASR system.

The dataset \textit{cleaned incremental transcripts with artificial noise} consists of the utterances from the dataset \textit{cleaned incremental transcripts} to these artificial noise were added with the approach described by \citet{SperberNW2017}.
We published the implementation\footnote{\url{https://github.com/msc42/NLP-tools/blob/master/noise_adder.py}} of this approach.
In this approach, random distributions are used to substitute, insert, and delete words.
We sampled the words for substitution and insertion based on acoustic similarity to the original input.
As vocabulary for the substitutions and insertions, we used the tokens of the utterances of the training dataset of the \textit{cleaned incremental transcripts} dataset and filled the vocabulary with the most frequent tokens not included in the used training dataset occurring in the source utterances of a subset of the OpenSubtitle corpus\footnote{based on \url{http://www.opensubtitles.org/}} \cite{Tiedemann2009} that is publicly available\footnote{\url{https://s3.amazonaws.com/opennmt-trainingdata/opensub\_qa\_en.tgz}} \cite{Senellart2017}.
We chose the position of the words to be substituted and deleted based on the length.
Shorter words are often more exposed to errors in ASR systems and therefore should be substituted and deleted in the artificial noise approach more frequently.
Since substitutions are more probable in ASR systems, we reflected this in the artificial noise generating by assigning substitutions a 5-times higher probability than insertions or deletions.
For the value of the hyperparameter \(\tau\) (the induced amount of noise), we used 0.08.

For the dataset \textit{human full transcripts}, we used the human transcribed transcripts given by the LDC CDs.
We mapped these utterances to the corresponding targets of the datasets based on the \textit{cleaned full transcripts} dataset.
The utterances are not cleaned and have some annotations like noise and repeated words.
The dataset \textit{human incremental transcripts}, \textit{human incremental transcripts with artificial noise}, and \textit{human full transcripts with artificial noise} were generated analogous to the described approaches before.

For the dataset \textit{automatic incremental transcripts}, we automatically transcribed the audio files from the LDC CDs with the ASR system Janus Recognition Toolkit (JRTk) \cite{NguyenMSZSW2017, NguyenSSW2018}.
This ASR system is used as an out-of-domain ASR system - there is no adaption for the ATIS utterances.
We used the incremental mode of the JRTk, which means that transcriptions are updated multiple times while transcribing.
It is not automatically possible to generate the partial output targets to the partial utterances, because the ASR system makes errors and it is impossible to map with 100\,\% accuracy automatically the wrong transcript \textit{to come up} to the correct transcript \textit{Tacoma}, for example.
We used a workaround: We measured the length of a partial transcript, searched the corresponding transcript of the \textit{human incremental transcripts} dataset that has the same length, and used the target of the found transcript.
If there were only shorter transcripts, the target of the full transcript was used.
This approach punishes insertions and deletions of the ASR system.
For the dataset \textit{automatic full transcripts}, we used the last transcript of the incremental transcripts of the ASR system for the user utterance and the full target of the corresponding utterance of the \textit{human full transcripts} dataset.
For the mentioned missing audio file, we used the human transcription of the corresponding LDC CD.

An arbitrary model without modifications, in this work the Transformer architecture, is trained with one of the described noisy datasets to have improved robustness compared to a model trained only with clean utterances.

\subsection{Evaluation metrics}
\label{sec:EvaluationMetrics}
We evaluated the quality of the models, trained with the different datasets, with the metric F\textsubscript{1}-score for which we used an adapted definition for the precision and the recall in this work and the metric intents accuracy.

The adapted definitions for the precision and the recall consider the order of the classes in the target sequence.
The intents and the intents parameters are the classes.
Intents parameters with the same slot tag are considered as different classes.
We call the F\textsubscript{1}-score calculated with the adapted definition of the precision and the recall \textbf{c}onsidering \textbf{o}rder \textbf{m}ultiple \textbf{c}lasses F\textsubscript{1}-score (CO-MC F\textsubscript{1}-score).
Order considering means that the predicted parameters have to be in the correct order in the target sequence.
In the target sequence
\begin{flushleft}\textit{atis\_flight fromloc.city\_name milwaukee toloc.city\_name orlando depart\_date.day\_name wednesday depart\_time.period\_of\_day evening or or depart\_date.day\_name thursday depart\_time.period\_of\_day morning}
\end{flushleft}
the order is important.
To calculate the true positives, we adapted the Levenshtein distance \cite{Levenshtein1966}.
The entities that are compared in this adapted Levenshtein distance are the classes.
The adapted Levenshtein distance is only changed by a match (incremented by one) and the maximum instead of the minimum function is used to select the best operation.
In Figure \ref{fig:LevenshteinDistance} the recursive definition of the adapted Levenshtein distance (ALD) is depicted.
Let \(r\) be the reference and \(h\) the hypothesis and \(|r|\) and \(|h|\) the number of classes of the reference or hyptothesis respectively and \(r_i\) and \(h_i\) the \(i\)-th class of the reference or hypothesis respectively.
\(L_{|h|, |r|}\) is the resultant adapted Levenshtein distance and the number of true positives.

\begin{figure}[H]

\begin{align*}
& ALD_{0,0} = 0 \\
& ALD_{i,0} = i, 1 \leq i \leq |h| \\
& ALD_{0,j} = j, 1 \leq j \leq |r| \\
& ALD_{i,j} = max \begin{cases}
               ALD_{i-1,j-1} + 1,~r_i = h_j \\
               ALD_{i-1,j-1},~r_i \neq h_j \\
               ALD_{i-1,j} \\
               ALD_{i,j-1} \\
\end{cases} \\
& ~~~~~ 1 \leq i \leq |h|, 1 \leq j \leq |r|
\end{align*}

\caption{adapted Levenshtein distance}
\label{fig:LevenshteinDistance}
\end{figure}

With this approach, the given example target has 7 instead of 9 true positives if the slot tokens of the intents parameters with the slot tag depart\_date.day\_name parameter are changed (in this case both parameters are considered as substitutions in the Levenshtein distance).
We counted all true positives for the different classes over the evaluated dataset and divided the counted true positives by the reference lengths of all targets for the recall and by the hypothesis lengths for the precision (micro-averaging).
The CO-MC F\textsubscript{1}-score is more strict than the vanilla F\textsubscript{1}-score because of the consideration of the order.

The metric intents accuracy considers all intents as whole.
That means the intents accuracy of one target is 100\,\% if the intents of the reference and the hypothesis are equivalent; otherwise, the intents accuracy is 0\,\%.

\subsection{Evaluation schemes}
\label{sec:EvaluationData}
We used for the evaluation of the models the model version of the epoch with the best CO-MC F\textsubscript{1}-score on the following validation datasets with only full utterances:
For the models trained with the datasets based on the \textit{cleaned full transcripts} dataset, we used the validation dataset of the \textit{cleaned full transcripts} dataset, for models trained with the datasets based on the \textit{human full transcripts} dataset, we used the validation dataset of the \textit{human full transcripts} dataset, and for models trained with the datasets based on the \textit{automatic incremental transcripts} dataset, we used the validation dataset of the \textit{automatic full transcribed} dataset.

We evaluated our models with our evaluation metrics in the following manner:
First, we evaluated the models with partial utterances that contain the first 100\,\%, 75\,\%, 50\,\%, and 25\,\% of the tokens of the full utterances.
The number of tokens is rounded off to the next integer and this number is called \(i\) in the following.
For evaluating with the cleaned and the human transcribed utterances, we used the first \(i\) tokens of the full utterances.
For evaluating with automatically transcribed utterances, we used the first utterance in the \textit{automatic incremental transcripts} dataset of the corresponding utterance that was equal than or greater than \(i\), because the ASR system did not produce partial utterances for all numbers less than the token length of the full utterance.
In the following, this evaluation scheme is called \textit{partial utterances processing}.

In addition, we evaluated our models with the metric intents accuracy in the following manner:
We predicted the intents incrementally and aborted the incremental processing once a certain confidence for the intents prediction was reached.
We used 95\,\%, 90\,\%, 85\,\%, and 80\,\% as confidence thresholds.
When the target confidence was never reached, the full utterance was used to predict the intents, even if the confidence of the full utterance was under the confidence threshold.
We used for those experiments the partial utterances successively for the cleaned and human transcribed utterances and the partial utterances successively of the automatically transcribed utterances.
In the automatically transcribed utterances, the last transcript is the full utterance.
In the following, this evaluation scheme is called \textit{confidence based processing}.

The models trained on the cleaned transcripts cannot be evaluated appropriately on the uncleaned transcripts, because the numbers are written in Arabic numerals in the cleaned transcripts and in words in the uncleaned transcripts.
The conversion is often ambiguous.
The same applies to the other direction.

\subsection{System Setup}
\label{sec:Setup}
We optimized the Transformer architecture for the validation dataset of the \textit{cleaned full transcripts} dataset.
The result of this optimization is a Transformer architecture with a model and inner size of 256, 4 layers, 4 heads, Adam \cite{KingmaB2015} with the noam learning rate decay scheme (used by \citet{VaswaniSPUJGKP2017} as learning rate decay scheme) as optimization algorithm, a dropout of 40\,\%, an attention, embedding, and residual dropout of each 20\,\% and a label smoothing of 15\,\%.
We used 64 utterances as batch size.
The vocabulary of a trained model contains all words of the training dataset with which it was trained.
We trained the models for 100 epochs.

\section{Results}
\label{sec:Results}

\subsection{Partial utterances processing}
In Tables \ref{tab:ResultsCleanedPart}, \ref{tab:ResultsHumanPart}, and \ref{tab:ResultsAutomaticPart}, the CO-MC F\textsubscript{1}-scores and the intents accuracies are depicted for the evaluation scheme \textit{partial hypothesis processing} for the cleaned, human transcribed, and automatically transcribed utterances respectively.

In the following, all percentage differences are absolute percentage differences.
The ranges refer to the smallest and biggest improvements on the CO-MC F\textsubscript{1}-score.
If no artificial noise is explicitly mentioned, the models without artificial noise are meant.

The models that were trained only with full utterances have better results evaluated on the full utterances than models trained with the partial and full utterances (in the range from 1.3\,\% to 3.24\,\%).
However, the models trained on the partial and full utterances have better results when they are evaluated on the first 75\,\% and 50\,\% of the tokens (in the range from 0.81\,\% to 4.39\,\%).
Evaluated on the utterances of the first 25\,\% of the tokens, there are even bigger improvements (in the range from 14.44\,\% to 47.91\,\%).
This means that our proposed training method improves the processing of partial utterances, especially if they are partial utterances produced incrementally at the beginning of the incremental processing of an utterance.
For an incremental processing capable NLU component, the best approach is to combine the two models.
The model trained on only full utterances is used for the full utterances and the model trained on the partial and full utterances is used for the incrementally produced partial utterances.

With the combination described above, the performance of the models trained with the automatically transcribed utterances decreased less compared to the models trained on the human transcribed utterances, evaluated on the human transcribed utterances (in the range from 0.13\,\% to 2.01\,\%) than the models trained with the human transcribed utterances decreased compared to the models trained on the automatically transcribed utterances, evaluated on the automatically transcribed utterances (in the range from 1.22\,\% to 4\,\%).
In our experiments, the result was consequently that it is better to train on noisier data.
This is especially the case on evaluating the full utterances.

We tried to simulate the noise of the automatically transcribed utterances with artificial noise.
We used again the same combination described above.
The performance of the models trained with the human transcribed utterances with artificial noise decreased less compared to the models trained on the human transcripts, evaluated on the human transcribed utterances (in the range from \mbox{-1.43\,\%} to 2.5\,\%) than the models trained with the human transcribed utterances decreased compared to the human transcribed utterances with artificial noise, on the automatically transcribed utterances (in the range from -1.06\,\% to 5.21\,\%).

\subsection{Confidence based processing}
In Tables \ref{tab:ResultsCleanedConfidence}, \ref{tab:ResultsHumanConfidence}, and \ref{tab:ResultsAutomaticConfidence}, the intents accuracies and the needed percentage of tokens on average are depicted for the evaluation scheme \textit{confidence based processing} for the cleaned, human transcribed, and automatically transcribed utterances respectively.

In the following, all percentage differences are absolute percentage differences.
The ranges refer to the smallest and biggest improvements on the intents accuracy metric.
If no artificial noise is explicitly mentioned, the models without artificial noise are meant.

The following statements apply to the incrementally trained models (the models trained only on the full utterances have only good results if they can use nearly the full utterances and therefore it makes no sense to use them for early predicting of intents).
It is better to train on the automatically transcribed utterances.
The decreasing is from 1.57\,\% to 2.58\,\% if they are evaluated on the human transcribed utterances, but they have an improvement from 2.58\,\% to 4.25\,\% if they are evaluated on the automatically transcribed utterances compared to the models trained on the human transcribed utterances.
The human transcribed utterances with artificial noise decrease by -1.46\,\% to 2.58\,\% if they are evaluated on the human transcribed utterances, but they have an improvement from 0.67\,\% to 3.69\,\% if they are evaluated on the automatically transcribed utterances compared to the models trained on the human transcribed utterances.

\begin{table*}
\centering

\begin{tabular}{ l | l | l | l | l }
training dataset & first 100\,\% & first 75\,\% & first 50\,\% & first 25\,\% \\
\hline
cleaned, full & 92.90 / 97.09 & 89.95 / 96.19 & 88.98 / 90.37 & 49.36 / 49.05 \\
cleaned, incremental & 91.60 / 96.75 & 94.20 / 94.85 & 93.37 / 92.05 & 83.15 / 79.73 \\
cleaned, incremental, art. noise & 91.97 / 96.19 & 94.65 / 94.85 & 93.22 / 91.83 & 81.75 / 78.61 \\
\end{tabular}

\caption{CO-MC F\textsubscript{1}-scores / intents accuracies of the first 100\,\%, 75\,\%, 50\,\%, and 25\,\% of the tokens of the utterances of the test dataset of the cleaned human transcribed full utterances}
\label{tab:ResultsCleanedPart}
\end{table*}

\begin{table*}
\centering

\begin{tabular}{ l | l | l | l | l }
training dataset & 95\,\% conf. & 90\,\% conf. & 85\,\% conf. & 80\,\% conf. \\
\hline
cleaned, full & 96.98 / 96.36 & 80.29 / 74.14 & 72.56 / 30.62 & 70.66 / 29.93 \\
cleaned, incremental & 96.42 / 97.27 & 95.97 / 92.16 & 92.83 / 34.74 & 89.47 / 30.56 \\
cleaned, incremental, art. noise & 95.86 / 99.02 & 95.41 / 92.20 & 91.71 / 33.12 & 86.56 / 24.54 \\
\end{tabular}

\caption{Intents accuracies / percentages of the used tokens for predicting the intents using the smallest partial utterance of the test dataset of the cleaned human transcribed incremental utterances for which the system has a confidence of more or equal than 95\,\%, 90\,\%, 85\,\%, and 80\,\%, if the confidence is not reached, the full utterance is used}
\label{tab:ResultsCleanedConfidence}
\end{table*}

\begin{table*}
\centering

\begin{tabular}{ l | l | l | l | l }
training dataset & first 100\,\% & first 75\,\% & first 50\,\% & first 25\,\% \\
\hline
human, full & 90.44 / 94.85 & 87.91 / 94.51 & 87.75 / 89.14 & 34.86 / 48.38 \\
human, full, art. noise & 87.94 / 95.30 & 85.77 / 94.74 & 89.51 / 91.27 & 67.71 / 68.65 \\
human, incremental & 88.57 / 94.40 & 90.58 / 93.62 & 91.51 / 90.59 & 82.77 / 79.06 \\
human, incremental, art. noise & 88.24 / 95.41 & 90.71 / 94.18 & 92.94 / 91.83 & 84.14 / 79.17 \\
automatic, full & 88.43 / 93.39 & 86.18 / 93.62 & 89.24 / 90.37 & 56.80 / 70.66 \\
automatic, incremental & 86.38 / 92.72 & 89.56 / 93.73 & 90.05 / 89.03 & 82.64 / 79.17 \\
\end{tabular}

\caption{CO-MC F\textsubscript{1}-scores / intents accuracies of the first 100\,\%, 75\,\%, 50\,\%, and 25\,\% of the tokens of the utterances of the test dataset of the human transcribed full utterances}
\label{tab:ResultsHumanPart}
\end{table*}

\begin{table*}
\centering

\begin{tabular}{ l | l | l | l | l }
training dataset & 95\,\% conf. & 90\,\% conf. & 85\,\% conf. & 80\,\% conf. \\
\hline
human, full & 94.51 / 96.51 & 90.82 / 85.59 & 77.60 / 32.52 & 76.37 / 30.22 \\
human, full, art. noise & 95.41 / 96.33 & 90.71 / 81.50 & 77.72 / 30.58 & 76.04 / 28.13 \\
human, incremental & 94.18 / 99.10 & 93.95 / 89.53 & 90.59 / 32.83 & 88.47 / 27.04 \\
human, incremental, art. noise & 95.18 / 97.85 & 95.41 / 92.65 & 91.60 / 32.78 & 85.89 / 24.90 \\
automatic, full & 88.58 / 91.51 & 88.35 / 83.19 & 76.82 / 30.85 & 75.36 / 28.79 \\
automatic, incremental & 92.61 / 99.63 & 92.16 / 93.38 & 88.35 / 35.41 & 85.89 / 30.33 \\
\end{tabular}

\caption{Intents accuracies / percentages of the used tokens for predicting the intents using the smallest partial utterance of the test dataset of the human transcribed incremental utterances for which the system has a confidence of more or equal than 95\,\%, 90\,\%, 85\,\%, and 80\,\%, if the confidence is not reached, the full utterance is used}
\label{tab:ResultsHumanConfidence}
\end{table*}

\begin{table*}
\centering

\begin{tabular}{ l | l | l | l | l }
training dataset & first 100\,\% & first 75\,\% & first 50\,\% & first 25\,\% \\
\hline
human, full & 83.87 / 91.49 & 80.26 / 91.04 & 83.13 / 85.78 & 42.06 / 51.74 \\
human, full, art. noise & 81.93 / 91.15 & 78.94 / 90.71 & 83.76 / 88.35 & 74.98 / 68.31 \\
human, incremental & 80.63 / 87.91 & 82.16 / 88.24 & 85.85 / 85.44 & 80.49 / 75.93 \\
human, incremental, art. noise & 82.93 / 91.04 & 83.16 / 89.70 & 88.13 / 88.02 & 83.75 / 77.27 \\
automatic, full & 87.14 / 93.39 & 82.06 / 92.61 & 85.15 / 90.03 & 70.05 / 72.45 \\
automatic, incremental & 84.62 / 92.27 & 84.90 / 91.71 & 87.07 / 88.35 & 84.49 / 79.73 \\
\end{tabular}

\caption{CO-MC F\textsubscript{1}-scores / intents accuracies of the first partial automatically transcribed utterances that have equal or more than the first 100\,\%, 75\,\%, 50\,\%, and 25\,\% of the tokens of the utterances of the test dataset of the automatically transcribed full utterances}
\label{tab:ResultsAutomaticPart}
\end{table*}

\begin{table*}
\centering

\begin{tabular}{ l | l | l | l | l }
training dataset & 95\,\% conf. & 90\,\% conf. & 85\,\% conf. & 80\,\% conf.\\
\hline
human, full & 91.04 / 97.63 & 87.46 / 88.24 & 78.84 / 41.52 & 76.82 / 38.41 \\
human, full, art. noise & 90.93 / 96.87 & 86.79 / 85.60 & 77.94 / 37.99 & 76.48 / 34.93 \\
human, incremental & 87.68 / 99.18 & 87.35 / 91.23 & 86.56 / 42.35 & 83.99 / 36.96 \\
human, incremental, art. noise & 90.59 / 98.29 & 90.37 / 93.58 & 88.47 / 40.33 & 82.98 / 31.53 \\
automatic, full & 88.24 / 93.41 & 87.91 / 86.65 & 80.40 / 38.74 & 78.50 / 35.70\\
automatic, incremental & 91.83 / 99.16 & 91.60 / 94.29 & 89.14 / 39.67 & 86.67 / 34.54 \\
\end{tabular}

\caption{Intents accuracies / percentages of the used tokens for predicting the intents using the first partial utterance of the test dataset of the automatically transcribed incremental utterances for which the system has a confidence of more or equal than 95\,\%, 90\,\%, 85\,\%, and 80\,\%, if the confidence is not reached, the full utterance is used}
\label{tab:ResultsAutomaticConfidence}
\end{table*}

\subsection{Computation time}
Since the partial utterances are fed successively in the Transformer architecture, the computation must be fast enough for the system to work off all partial utterances without latency.
On a notebook with an Intel Core i5-8250U CPU - all computations were done only on the CPU and we limited the usage to one thread (with the app taskset) so other component like the ASR system can run on the same system - it took 310 milliseconds to compute the longest utterance (46 tokens) of the cleaned utterances and 293 milliseconds to compute the utterance (38 tokens) with the longest target sequence (41 tokens - one intent with 17 parameters) of the cleaned utterances.
We processed continually both utterances for 15 minutes and selected for both utterances the run with the maximum computation time.
The model was the model trained with the cleaned full utterances.
This means that it is possible to process an utterance after every word because a normal user needs on average more than these measured times to utter a word or type a word with a keyboard.

\section{Conclusions and Further Work}
\label{sec:Conlusion}
In this work, we report that the best approach for an incremental processing capable NLU component is to mix models.
A model trained on partial and full utterances should be used for processing partial utterances and a model trained only on full utterances for processing full utterances.
In particular, the improvements are for the first incrementally produced utterances, which contain only a small number of tokens, high if the model is not only trained on full utterances.

Evaluated on the noisy human and even noisier automatically transcribed utterances, we got better results with the models trained with the human transcribed utterances with artificial noise and the models trained with the automatically transcribed utterances.
This is especially the case when evaluating the full utterances.
A reason for this could be that the partial utterances can be already considered as noisier utterances.

The short computation time of the processing of an utterance makes it possible to use the incremental processing for spoken and written utterances.

In future work, it has to be evaluated whether our results are also valid for other architectures and other datasets.
A balanced version of the ATIS datasets can also be seen as another dataset.

We got better performance with artificial noise.
However, the results could be improved by optimizing the hyperparameter of the artificial noise generator.

In this work, we researched the performance using incremental utterances.
There should be research on how the results of the incremental processing can be separated into subactions and how much this can accelerate the processing of actions in real-world scenarios.

In future work not only the acceleration, but also other benefits of the incremental processing, like using semantic information for improving the backchannel, could be researched.

\section*{Acknowledgement}
This work has been conducted in the SecondHands project which has received funding from the European Union's Horizon 2020 Research and Innovation programme (call:H2020- ICT-2014-1, RIA) under grant agreement No 643950.

\bibliography{bibliography}
\bibliographystyle{acl_natbib_nourl}

\appendix
\section{Supplemental Material}
\label{sec:supplemental}
\begin{table}[H]
\centering

\begin{tabular}{ l | l | l | p{3.4cm} }
train & valid & test & intent(s) \\ \hline
73.89 & 71.4 & 70.77 & atis\_flight \\
8.6 & 7.6 & 5.38 & atis\_airfare \\
5.14 & 5.0 & 4.03 & atis\_ground\_service \\
3.1 & 3.6 & 4.26 & atis\_airline \\
2.9 & 3.4 & 3.7 & atis\_abbreviation \\
1.56 & 2.2 & 1.01 & atis\_aircraft \\
1.0 & 1.8 & 0.11 & atis\_flight\_time \\
0.92 & 2.0 & 0.34 & atis\_quantity \\
0.42 & 0.4 & 1.34 & atis\_flight\#atis\_airfare \\
0.4 & 0.2 & 0.67 & atis\_city \\
0.38 & 0.6 & 1.12 & atis\_distance \\
0.38 & 0.6 & 2.02 & atis\_airport \\
0.33 & 0.6 & 0.78 & atis\_ground\_fare \\
0.33 & 0.2 & 2.35 & atis\_capacity \\
0.27 & 0 & 8 & atis\_flight\_no \\
0.13 & 0 & 0.67 & atis\_meal \\
0.11 & 0.2 & 0 & atis\_restriction \\
0.04 & 0 & 0 & atis\_airline\# atis\_flight\_no \\
0.02 & 0 & 0 & atis\_ground\_service\# atis\_ground\_fare \\
0.02 & 0 & 0 & atis\_cheapest \\
0.02 & 0 & 0 & atis\_aircraft\#atis\_flight\# atis\_flight\_no \\
0 & 0.2 & 0 & atis\_airfare\# atis\_flight\_time \\
0 & 0 & 0.22 & atis\_day\_name \\
0 & 0 & 0.11 & atis\_flight\#atis\_airline \\
0 & 0 & 0.11 & atis\_airfare\#atis\_flight \\
0 & 0 & 0.11 & atis\_flight\_no\# atis\_airline \\
\end{tabular}

\caption{intents distribution (in percent) of the ATIS utterances used in this work}
\label{fig:Intents_distribution}
\end{table}

\end{document}